\documentclass[letterpaper]{article}
\usepackage{aaai2026}
\usepackage{times}
\usepackage{helvet}
\usepackage{courier}
\usepackage[hyphens]{url}
\usepackage{graphicx}
\usepackage{natbib}
\usepackage{caption}
\usepackage{booktabs}
\usepackage{amsmath}
\usepackage{xcolor}
\usepackage{siunitx}

\title{Structure First, Reason Next: Enhancing a Large Language Model using Knowledge Graph for Numerical Reasoning in Financial Documents}

\author{
  Aryan Mishra\textsuperscript{1},
  Akash Anil\textsuperscript{1}
}

\affiliations{
  \textsuperscript{1}Department of Data Science and Engineering,\\
  Indian Institute of Science Education and Research Bhopal, Madhya Pradesh, India
}

\begin{document}
\maketitle

\begin{abstract}
Numerical reasoning is an important task in the analysis of financial documents. It helps in understanding and performing numerical predictions with logical conclusions for the given query seeking answers from financial texts. 
Recently, Large Language Models (LLMs) have shown promising results in multiple Question-Answering (Q-A) systems with the capability of logical reasoning. As documents related to finance often consist of long and complex financial contexts, LLMs appear well-suited for building high-quality automated financial question-answering systems. However, LLMs often face challenges in accurately processing the various numbers within financial reports. 

Extracting numerical data from unstructured text and semi-structured tables, and reliably performing accurate calculations, remains a significant bottleneck for numerical reasoning in most state-of-the-art LLMs. Recent studies have shown that structured data augmentations, such as Knowledge Graphs (KGs), have notably improved the predictions of LLMs along with logical explanations. Thus, it is an important requirement to consider inherent structured information in financial reports while using LLMs for various financial analytics. 

This paper proposes a framework to incorporate structured information using KGs along with LLM predictions for numerical reasoning tasks. The KGs are extracted using a proposed schema inherently from the document under processing. We evaluated our proposed framework over the benchmark data FinQA, using an open-source LLM, namely Llama 3.1 8B Instruct. We observed that the proposed framework improved execution accuracy by approximately 12\% relative to the vanilla LLM.

\end{abstract}

\section{Introduction}

Numerical Reasoning in financial data refers to the analysis and interpretation of quantitative information such as revenue figures, ratios, market indicators, or statistical trends present in the financial reports~\cite{chen2021finqa}. Although LLMs demonstrate promising reasoning capabilities in various domains, they often show limited performance when subjected to financial reasoning~\cite{qian2025fino1,liu2025finr1}. The limited capability of LLMs is mainly due to the quantitative characteristics of financial data incorporating multiple paragraphs and tables with numbers, which makes it harder to exploit the inherent context~\cite{nie2024surveylargelanguagemodels}. 

A majority of the LLMs show promising reasoning capability yet they are often limited in some of the specialized domains (e.g., finance, healthcare) because they primarily learn from unstructured text data, relying on statistical co-occurrences rather than inherent relational characteristics~\cite{tan2024structxenhancinglargelanguage}. To address this limitation, recently some of the studies integrate structured information such as Knowledge Graphs (KGs) for enhancing the reasoning abilities of LLMs~\cite{sun2024pyramiddrivenalignmentpyramidprinciple}. KGs provide semantic relationships and factual grounding and thus found to be helpful in improving reasoning performance in many domains~\cite{wu2024thinkingknowledgegraphsenhancing}. However, to the best of our knowledge, none of these studies explicitly address numerical reasoning over financial data while capturing inherent structural aspects.

\begin{figure*}[t!]
\centering
\resizebox{1\textwidth}{0.53\textheight}{%
    \includegraphics{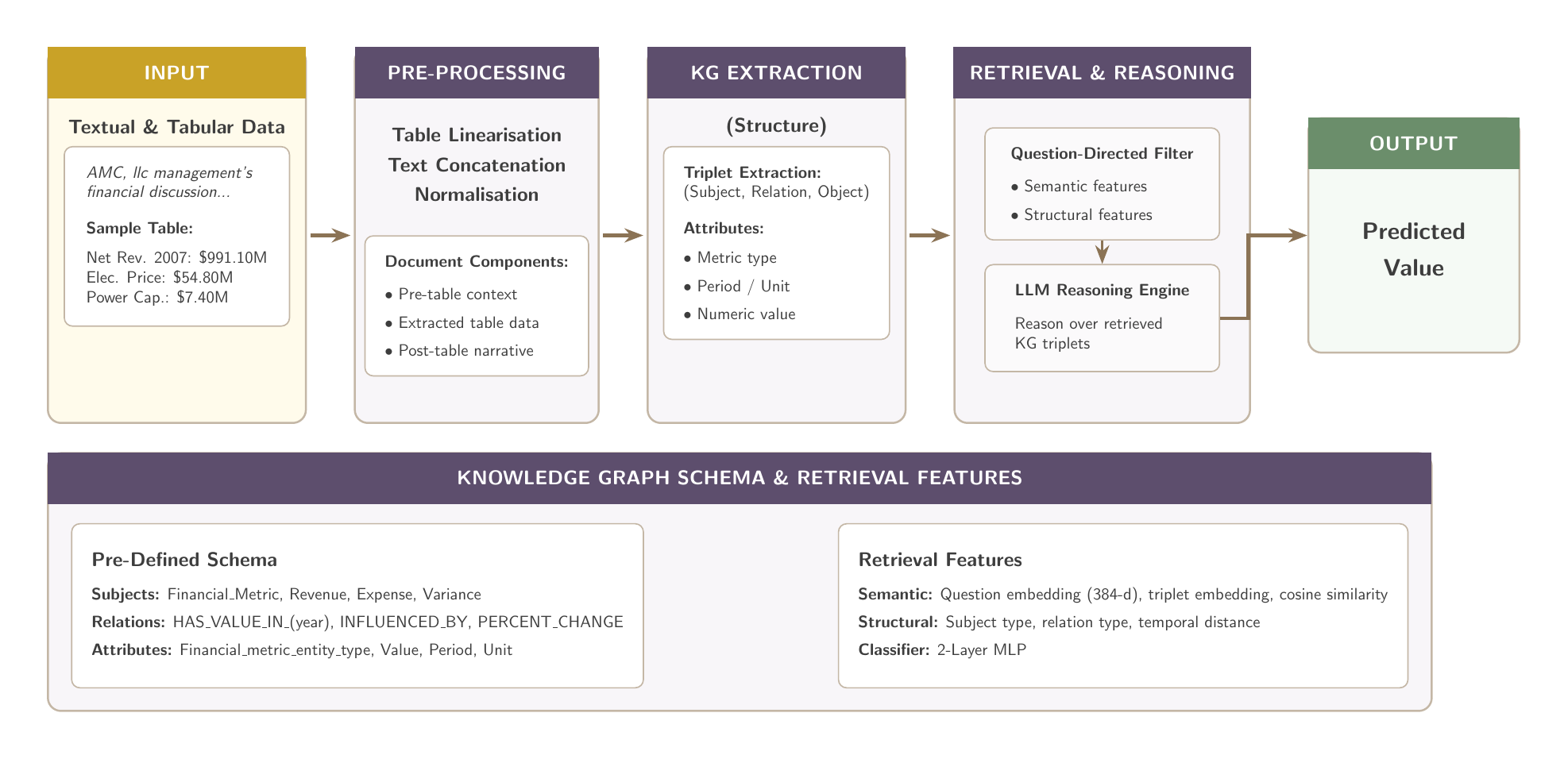}%
}
\caption{end-to-end pipeline for numerical reasoning in financial data with inherent knowledge graph and LLMs.}
\label{fig:pipeline}
\end{figure*}

To bridge the gap in exploiting structural information in numerical reasoning for finance data, we propose a novel framework that uses inherent KG extracted using predefined schema and an open-source LLM. Figure ~\ref{fig:pipeline} presents an end-to-end pipeline for the proposed framework. Our framework (i) preprocesses documents (including table linearization), (ii) constructs knowledge graphs using predefined schema, financial entities, and temporal relationships using few-shot prompting, (iii) performs lightweight retrieval combining semantic and structural features, and (iv) reasons using any LLM exploiting the structured input for predicting the output.


We evaluate the proposed framework\footnote{Code will be released publicly upon publication.} using Llama 3.1 8B Instruct (Llama)\cite{grattafiori2024llama3herdmodels} on state-of-the-art financial reasoning benchmark namely FinQA~\cite{chen2021finqa}. Further, we systematically compare the performance of the proposed framework to the open-source Llama model. It is evident that the proposed framework using KGs considerably enhanced the performance of vanilla Llama model. 

To summarize, the main contributions of this paper are:
\begin{enumerate}
    \item Study numerical reasoning in financial data using LLMs exploiting structural information in the form of Knowledge Graphs.
    \item Build an end-to-end pipeline for numerical reasoning capable of preprocessing, extracting KGs, retrieval, and enhanced reasoning. 
    \item Systematically compare the results against suitable baseline. 
\end{enumerate}

Rest of the paper is organized as follows. Section~\ref{sec:related_study} presents related studies which is followed by a detailed discussion on the proposed framework in Section~\ref{sec:proposed_framework}. In Section~\ref{sec:experimental_setup}, we discuss the experimental setup and Section~\ref{sec:result_analysis} discusses the performance of proposed framework compared with the baseline. Section~\ref{sec:conclusion} concludes the paper. Further, Section~\ref{sec:limitations} presents a limitation overview.

\section{Related Work}
\label{sec:related_study}
The FinQA benchmark \cite{chen2021finqa} formalized numerical question answering over financial documents and shown various analytics. In a similar direction,  authors of APOLLO in~\cite{sun2024apollo} introduced number-aware negative sampling and~\cite{li2023dyrren} incorporates dynamic re-ranking in financial reasoning.    
Recently,~\cite{qian2025fino1} proposed a domain specific finetuning framework that enhances reasoning capability. Critically,\cite{qian2025fino1} observe that even heavily fine-tuned models (FinR1~\cite{liu2025finr1}, Dianjin-R1~\cite{zhu2025dianjinr1evaluatingenhancingfinancial}) exhibit performance 
degradation on longer and complex documents, falling below base models.
Recently,~\cite{srivastava2024evaluating} categorized FinQA queries into 
arithmetic operations such as SUM, DIFFERENCE, RATIO, CHANGE\_RATIO, highlighting various challenges in multi-step domain-specific reasoning. These studies highlight major bottlenecks in numerical reasoning over financial text often considering longer and tabular structure. 


To improve the reasoning using LLMs, Retrieval Augmented Generation (RAG) has garnered attention. RAG-based frameworks ground LLM outputs with external evidence~\cite{lewis2020retrieval} to predict a more curated output and thus enhances the performance of LLMs. HybridRAG~\cite{sarmah2024hybridrag} combines vector 
and graph retrieval, though concatenating contexts increases cognitive load and in financial contexts, naive text chunking disrupts numerical links. Thus, our proposed framework attempts to mitigate this issue by fusing semantic and structural features in a lightweight 
retriever, optimizing retrieval without context concatenation overhead.

In the recent past, SubgraphRAG~\cite{li2025subgraphrag} demonstrates that lightweight MLPs with engineered graph features outperform complex graph neural networks for retrieval. In a similar direction, we aim to adapt curated domain-specific attributes such as temporal distances and entity types easily derived from FinQA dataset.

\section{Proposed Framework}
\label{sec:proposed_framework}
Figure~\ref{fig:pipeline} presents the proposed framework comprising of three main steps. After getting the input, the first step executes data preprocessing by table linearization and text concatenation with normalization for providing a uniform input text. We linearize tables following prior work on financial QA \cite{chen2021finqa}. Further, step two automatically extracts KG triplets from the input text provided from step one using a predefined schema proposed for financial reasoning. In step three, the framework filters the unwanted triplets and performs the reasoning task using preferred LLM model.

\subsection{Step 1: Document Preprocessing}
Financial documents often contain hybrid data, i.e., narrative texts and semi-structured tables. To enable uniform processing, we linearize tables using template-based conversion shown below:

\begin{small}
\begin{verbatim}
Original Table   Linearized Text
Year | Revenue    
                   
2020 | $100M   → "For 2020, revenue is $100M.

2021 | $120M   →  For 2021, revenue is $120M."
\end{verbatim}
\end{small}

Text Linearization enables text-based processing though loses explicit 
structural temporal relationships, and thus it might be the case that entity types are not differentiated which makes the numerical formats inconsistent. This inconsistency might be alleviated using the KG construction preserving temporal relationships.

\subsection{Step 2: Knowledge Graph Construction}
Knowledge Graphs are one of the key features in our framework. Thus, for constructing the KGs, we leverage the financial context understanding of LLMs and follow a standard schema for generating triplets. We aim to represent the information present in the text in unambiguous format and reduce the potential extraction errors. The inherent temporal and entity relational features are also preserved in this format. The structured triplets enable better processing and understanding of the multi-hop relations present across the document.
The schema for such KGs is focused on the numerical and temporal facts. 
The proposed schema for KG is given below:

\paragraph{Financial Domain Schema}

\begin{small}
\begin{verbatim}
Schema  = (subject, relation, object, 
            {financial_metric_entity_type,  
              company,period,value, unit})

Example:
subject: "NET_REVENUE:Entergy"
relation: "HAS_VALUE_IN_2015"
object: "5829 million USD"
financial_metric_entity_type: "NET_REVENUE"
company: "Entergy"
period: "2015"
value: "5829"
unit: "million USD"
\end{verbatim}
\end{small}

Now, for generating triplets across documents, we use few-shot-based prompts designed as below:

\paragraph{Prompt Engineering}
We provide comprehensive extraction rules via natural language prompt (key excerpts):

\begin{small}
\begin{verbatim}
Extract financial facts containing:
1. Detailed metric (NET_REVENUE, 
   OPERATING_EXPENSES, etc.)
2. Numerical value with units
3. Temporal qualifier (infer from context)

RULES:
- Use EXACT TEXT (no paraphrasing)
- Standardize periods: "2007", "AS_OF_2010",
  "2007-Q4", "AFTER_2015"
- Extract liberally for coverage
- Multiple periods → separate extractions

ATTRIBUTE REQUIREMENTS:
- SUBJECT: "METRIC:Company" or "METRIC"
- RELATION: "HAS_VALUE_IN_YYYY" variants
- OBJECT: "value unit" (human-readable)
- PERIOD: Standardized format
\end{verbatim}
\end{small}

\subsection{Step 3: Filtering Triplets and Reasoning}
Similar to SubgraphRAG \cite{li2025subgraphrag}, we integrate semantic embeddings and KG embeddings to a Multi-Layer Perceptron (MLP) classifier. The MLP classifier outputs relevant triplets and works as a filter. Further, any preferred LLM may be used for numerical reasoning task with the filtered triplets.

\section{Experimental Setups}
\label{sec:experimental_setup}
In this section, we discuss the experimental setups including dataset analysis, evaluation, and baseline selection.  

\subsection{Dataset}
In this paper we use the FinQA benchmark as the dataset. FinQA consists of 6251 training examples, 883 validation examples, and 1147 test examples. We select FinQA as it is one of the standard benchmarks used in multiple studies on financial reasoning~\cite{zhu2025dianjinr1evaluatingenhancingfinancial}, ~\cite{liu2025finr1}, ~\cite{srivastava2024evaluating}. 

In particular to this paper, the schema for KG extraction is based on FinQA attributes. With little modification, it can be applied to other financial reasoning datasets.

\subsection{Hyper-parameters to LLM (i.e., Llama) and MLP classifier}
To generate KGs based on the above schema, we use open-source LLM, namely Llama 3.1 8B Instruct (Llama). We chose Llama because it is freely available and requires only approximately 16 GB GPU RAM. Furthermore, we use the following hyper-parameters for Llama:
(i) Model: Llama 3.1 8B Instruct, (ii) Temperature: 0.2, (iii) Maximum Number of Tokens: 2048.

For MLP classifier, we used two layers and semantic features include question embedding, triplet embedding, and cosine similarity (question embedding, triplet embedding). Binary cross-entropy was used as a loss function.

\subsection{Evaluation Setup and Baselines}
Similar to~\cite{liu2025finr1} and~\cite{qian2025fino1}, we use execution accuracy as a performance metric. Furthermore, we follow the standard way similar to~\cite{zhu2025judgelmfinetunedlargelanguage} and choose Gemini 2.5 Pro as a Judge to evaluate.
Gemini 2.5 Pro (gemini-2.5-pro) as judge, evaluates the semantic equivalence with temperature 0.0. It accounts for the format differences (e.g., 20\% = 0.20), minor rounding (e.g., ±1\%), unit variations (e.g., \$1.2M = \$1,200,000), and semantic equivalence (e.g., 20\% increase = grew by 20\%). We chose Gemini as a judge because it was freely available unlike GPT-4o which have been considered by some of the recent works in this direction ~\cite{qian2025fino1}.

To compare the performance by the proposed framework, we use the vanilla Llama itself. We could not directly compare with the baselines in ~\cite{qian2025fino1} as they used proprietary version of GPT-4o for judgement and thus not comparable with our work.

\section{Result and Analysis}
\label{sec:result_analysis}
Table~\ref{tab:main} presents the main results for our framework using KG with Llama model. Using KGs yields a +6.41 percentage-point absolute improvement (51.93\% to 58.34\%), which corresponds to approximately a 12.3\% relative improvement in execution accuracy. This result confirms the capability of leveraging structured information for the financial reasoning.

\begin{table}[h!]
\centering
\small
\begin{tabular}{@{}p{0.52\columnwidth}S[table-format=2.2]S[table-format=+1.2]@{}}
\toprule
Method & {Acc. (\%)} \\
\midrule
Llama (vanilla) & 51.93\%  \\
Llama $+$ KG & 58.34\% &  \\
\bottomrule
\end{tabular}
\caption{ Execution Accuracy by Llama and Llama $+$ KG: For judging the predictions Gemini 2.5 Pro has been used in both of the settings.}
\label{tab:main}
\end{table}

\subsection{Why KG Structure Helps in Numerical Reasoning?}
We now analyze the limitations of LLMs using only the text inputs using the following three aspects:

\begin{itemize}
    \item \textbf{Temporal Disambiguation:} 
    Text inputs often contain multiple dates, and a semantic retrieval model based purely on similarity may fail to distinguish between queries such as ``2020 revenue'' and ``2020 expenses.'' 
    In contrast, a KG-based approach explicitly encodes structured attributes such as \texttt{period="2020"} and \texttt{financial\_metric\_type="REVENUE"}, therefore enabling precise filtering and accurate retrieval.

    \item \textbf{Numerical Precision:} In text-only inputs, extracting precise numbers from long documents is difficult; KGs preserve hierarchical and contextual information and can improve numerical accuracy.

    \item 
\textbf{Multi-hop Requirements:} Many related facts are usually separated by various paragraphs in financial texts. Using a KG-based solution, triplets with the same entity\_type groups related triplets and improves multi-hop retrieval and reasoning.

\end{itemize}

\section{Conclusion}
\label{sec:conclusion}
This paper studies the effects of augmenting knowledge graphs for numerical reasoning in financial dataset. This work is motivated by the past successes of the Retrieval Augmented Generation (RAG) using structured information such as knowledge graphs. Further, we notice that there is limited prior work incorporating the inherent and natural relational structure of the financial texts. This paper at first proposes an end-to-end pipeline that can be adapted using any large language model along with harnessing the structural properties of the texts. The proposed framework exploits a predefined KG schema, which can be easily updated for various types of financial datasets. 

With systematic experiments and suitable baseline we found that using structural information (e.g., KG) of the financial text improves the prediction and shows a better reasoning capability.

\section{Limitations}
\label{sec:limitations}
This work has some limitations at present, which are discussed below:

\begin{itemize}
    \item \textbf{Dataset:} We considered only a single standard benchmark dataset in financial reasoning, namely FinQA. However, there are some more available benchmarks and we intend to explore our proposed framework over these datasets in future.  
    \item \textbf{LLM:} Due to resource constraints and proprietary solutions, we could not use multiple recently proposed LLMs. We further intend to consider more open-source LLMs and verify the effectiveness of structural characteristics in financial reasoning task.
    \item \textbf{KG Schema:} Although the proposed KG schema can be easily updated, we find that multiple datasets have variance in terms of many entity types and relationship types. This may be a bottleneck when the proposed schema is used for some of the localized financial datasets.
    \item \textbf{Baseline:} We noticed that many of the previous studies on numerical reasoning with financial dataset consider proprietary LLMs. As our research is limited to using the open-source LLMs, a majority of the past works could not be compared. Further, we were not able to easily adapt the available works to have a comparison, as their implementations were not open due to the usage of proprietary LLMs such as GPT-4o for estimating execution accuracy.   
\end{itemize}

\bibliography{aaai2026}

\end{document}